\documentclass[journal]{IEEEtran}

\usepackage{graphicx}
\usepackage{amssymb}
\usepackage{booktabs}
\usepackage{multirow}
\usepackage{caption}
\usepackage{color,xcolor}
\usepackage{algorithm}
\usepackage{algorithmic}
\usepackage{animate}

\usepackage{bm}
\usepackage{amsmath}

\usepackage[pagebackref=true,breaklinks=true,colorlinks,bookmarks=false]{hyperref}

\ifCLASSOPTIONcompsoc
  \usepackage[nocompress]{cite}
\else
  \usepackage{cite}
\fi

\begin{document}
\title{Cross-Modality Earth Mover’s Distance for \\  Visible Thermal Person Re-Identification}

\author{Yongguo Ling, 
        Zhun Zhong, 
        Donglin Cao, 
        Zhiming Luo, 
        Yaojin Lin, 
        Shaozi Li, 
        Nicu Sebe
        
}

\markboth{Journal of \LaTeX\ Class Files,~Vol.~14, No.~8, August~2015}%
{Shell \MakeLowercase{\textit{et al.}}: Bare Demo of IEEEtran.cls for Computer Society Journals}

\IEEEtitleabstractindextext{%
\begin{abstract}
Visible thermal person re-identification (VT-ReID) suffers from the inter-modality discrepancy and intra-identity variations. Distribution alignment is a popular solution for VT-ReID, which, however, is usually restricted to the influence of the intra-identity variations.
In this paper, we propose the Cross-Modality Earth Mover's Distance (CM-EMD) that can alleviate the impact of the intra-identity variations during modality alignment. CM-EMD selects an optimal transport strategy and assigns high weights to pairs that have a smaller intra-identity variation. In this manner, the model will focus on reducing the inter-modality discrepancy while paying less attention to intra-identity variations, leading to a more effective modality alignment. Moreover, we introduce two techniques to improve the advantage of CM-EMD. First, the Cross-Modality Discrimination Learning (CM-DL) is designed to overcome the discrimination degradation problem caused by modality alignment. By reducing the ratio between intra-identity and inter-identity variances, CM-DL leads the model to learn more discriminative representations. Second, we construct the Multi-Granularity Structure (MGS), enabling us to align modalities from both coarse- and fine-grained levels with the proposed CM-EMD. Extensive experiments show the benefits of the proposed CM-EMD and its auxiliary techniques (CM-DL and MGS). Our method achieves state-of-the-art performance on two VT-ReID benchmarks.
\end{abstract}

}

\maketitle

\IEEEdisplaynontitleabstractindextext

\IEEEpeerreviewmaketitle

\section{Introduction}
\label{sec:introduction}

\IEEEPARstart{P}{erson} re-identification (ReID) is critical in a safety surveillance system aiming at matching a query of interest from a set of gallery images captured by non-overlapping cameras. Traditional ReID~\cite{Zhong_2018_CVPR,Cheng_2016_CVPR} assumes that the images are collected by RGB cameras. However, the appearance characteristics of visible (RGB) images are largely relying on the illumination conditions and will be highly influenced by the poor illumination (\textit{e.g.,} night-time). To handle the night-time context, the ReID community has captured night-time images by thermal cameras and proposed a task called visible thermal person re-identification (VT-ReID)~\cite{ye2019bi}. This task is a cross-modality retrieval problem, aiming to search a query of one modality (\textit{e.g.}, visible) from the gallery of another modality (\textit{e.g.}, thermal). VT-ReID is more challenging than the traditional ReID, because we need to jointly overcome the inter-modality discrepancy caused by visible and thermal cameras and the intra-identity variations caused by different factors (such as view, pose and background).

\begin{figure}[!t]
\centering
  \includegraphics[width=0.95\linewidth]{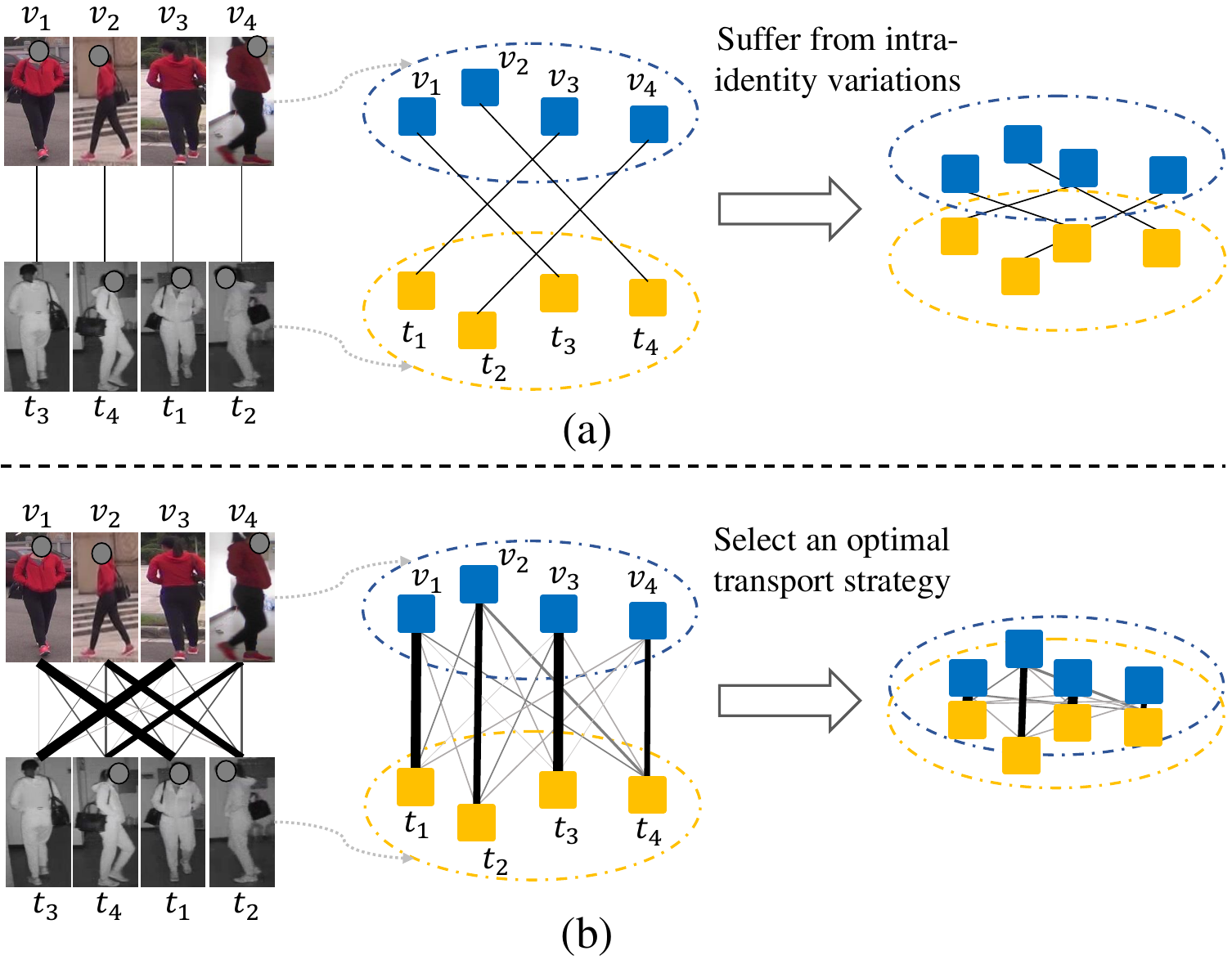}
  \vspace{-.05in}
  \caption{(a) Traditional distribution alignment. The randomly selected pairs usually have high intra-identity variations, hindering the reduction of inter-modality discrepancy. (b) Distribution alignment with Cross-Modality Earth Mover’s Distance (CM-EMD). Our CM-EMD assigns large weights to pairs that have less intra-identity variations, leading the model mainly focus on reducing the modality discrepancy rather than intra-identity variations.}
  \label{fig challenge}
\end{figure}

Distribution alignment~\cite{pu2020dual,hao2019dual,dai2018cross} is a popular and effective solution for VT-ReID, which aims to reduce the distribution gap between visible and thermal modalities. However, the existence of intra-identity variations may interfere with the optimization of distribution alignment. For example, given a training mini-batch, the randomly selected cross-modality pairs of the same identity commonly have a large variation in pose, view and background (Figure~\ref{fig challenge}~(a)). In this context, when the cross-modality discrepancy is reduced to a certain extent and is not dominant, the model will turn to decrease the intra-identity variations that are not suitable for optimizing by the distribution alignment function. Intuitively, if there are only cross-modality differences between all the selected cross-modality pairs, the model can always focus on reducing the modality gap to obtain a better modality alignment result. However, it is hard to achieve the above situation, since, in VT-ReID, the intra-identity variations always exist and such pairs are hard to obtain using only the identity annotations.

In this paper, we propose a novel distribution alignment approach for VT-ReID, called Cross-Modality  Earth Mover’s Distance (CM-EMD), which can largely mitigate the impact of the intra-identity variations during modality alignment. Specifically, given a mini-batch, CM-EMD leverages EMD~\cite{rubner1998emdmetric} to select an optimal transport strategy across two modalities, which assigns a large transport weight between two cross-modality samples that have a smaller intra-identity variation (Figure~\ref{fig challenge} (b)). In this way, the model can focus on reducing the modality discrepancy instead of the intra-identity variation, leading to an effective modality alignment. We also introduce two techniques to facilitate the effectiveness of CM-EMD. On the one hand, since CM-EMD mainly focuses on aligning the cross-modality distribution, this will inevitably degrade the discrimination ability of the representation. To overcome this issue, we present a Cross-Modality Discrimination Learning (CM-DL), which can improve the discrimination by reducing the ratio between the intra-identity and the inter-identity variances. In addition, knowing that local information is important for Re-ID~\cite{sun2018beyond}, we further introduce a Multi-Granularity Structure  (MGS) to perform finer modality alignment. In our MGS, we first extract both part-based local features and global features and then apply CM-EMD on them individually, enabling us to decrease the modality gap on both coarse-grained and fine-grained feature levels. In summary, the contributions of this paper are:
\begin{itemize}
    \item We introduce the Cross-Modality Earth Mover’s  Distance (CM-EMD) for VT-ReID, which can alleviate the negative effect caused by the intra-identity variations and effectively bridge the modality gap.
    \item We propose the Cross-Modality  Discrimination Learning (CM-DL), effectively overcoming the discrimination degradation problem raised by CM-EMD.
    \item We design the Multi-Granularity Structure (MGS), versatilely reducing the modality gap with CM-EMD.
\end{itemize}
Experiments on two VT-ReID datasets verify the advantages of the proposed CM-EMD, CM-DL and MGS, and demonstrate the superior performance of our method over state-of-the-art methods.

\section{Related Work}

Visible-Thermal person re-identification (VT-ReID) aims at matching a thermal / visible query person from a visible / thermal gallery. Generally, existing methods of VT-ReID can be mainly divided into four groups, \textit{i.e.}, feature extraction based methods, metric learning based methods, distribution alignment based methods, and image generation based methods.

\textbf{Feature Extractor based methods} mainly focus on designing a cross-modality network to extract modality-invariant and discriminative representation~\cite{wu2017rgb,ye2018visible,liu2020enhancing,yang2020mining,lu2020cross,ye2020dynamic,jia2020similarity,chen2021neural,tian2021farewell}.
Wu et al.~\cite{wu2017rgb} first consider the visible-thermal problem in ReID and introduce a visible-thermal dataset (SYSU-MM01). In addition, they inject a deep zero-padding approach into the network for evolving domain-specific nodes. Ye et al.~\cite{ye2018visible} introduce a two-stream network to handle the two modalities respectively and a dual-constrained top-ranking loss for learning robust cross-modality embedding.
However, Yang et al. \cite{yang2020mining} point out that the training of a two-stream model may distort the distribution of unseen classes. To tackle this issue, a bi-directional random walk scheme is proposed to discover more reliable cross-modality feature distributions.
Jia et al.~\cite{jia2020similarity} design a similarity inference metric (SIM) to leverage the similarities between intra-modality samples to help reduce cross-modality gap.
Lu et al.~\cite{lu2020cross} introduce a cross-modality feature transfer algorithm which can effectively extract the modality-shared information and modality-specific characteristic in the feature space. To extract more fine-grained features, Ye et al.~\cite{ye2020dynamic} develop a dynamic dual-attentive aggregation module by mining both intra-modality part-level and cross-modality graph-level contextual. Recently, Chen et al.~\cite{chen2021neural} propose Neural Feature Search (NFS) to automate select identity-related feature for matching. Tian et al.~\cite{tian2021farewell} introduce Variational Self-Distillation (VSD) to fitting the mutual information and reducing view and irrelevant information, learning more robustness representation.
These approaches design different feature extractors to extract either global-level feature or part-level feature. However, they ignore jointly considered these two aspects and studied their mutual benefits. Therefore, we devise a framework to extract and combine global- and part-level features, which can carry more discriminative content.

\textbf{Metric learning based methods} are proposed to learn an embedding space by explicitly enforcing the intra-class samples of two modalities close to each other~\cite{ye2018hierarchical,hao2019hsme,feng2019learning,ling2020class}. 
Ye et al.~\cite{ye2018hierarchical} introduce a hierarchical inter-modality metric learning to learn cross-modality embedding. Hao et al.~\cite{hao2019hsme} map the features to a hyper-sphere manifold which can effectively reduce the intra-modality variations and cross-modality variations with joint classification and identification constraints. In addition,  Ling et al.~\cite{ling2020class} introduce a center-guided loss function by formulating learning constraints among class centers and instances, which can effectively reduce inter- and intra- modalities discrepancy.  
Different from these approaches, this paper exploit  the Cross-modality Earth Mover's Distance (cmEMD) to reduce cross-modality discrepancy.

\textbf{Distribution alignment based methods} are mainly proposed to decrease the distribution divergence to smooth the inter-modality discrepancy and gain modality invariant feature~\cite{dai2018cross,hao2019dual,pu2020dual,wu2021discover}. 
Dai et al.~\cite{dai2018cross} introduce a generative adversarial network to design a modality discriminator to learn discriminative feature. Hao et al.~\cite{hao2019dual} propose dual-alignment to reduce the modality gap in the concept of spatial and modality by extracting part-level features and learning with distribution and correlation losses. Pu et al.~\cite{pu2020dual} propose a dual Gaussian-based variational auto-encoder (DG-VAE), which allows model to explore unobserved data enforce both modalities to follow the true distribution. In order to explore nuances information, Wu et al.~\cite{wu2021discover} introduce MPANet to align modality and extract discriminative features by a modality alleviation module and a pattern alignment module.
These approaches can alleviate the large inter-modality discrepancy. Nevertheless, the intra-modality variations may limit the distribution alignment of two modalities. Therefore, we propose the Cross-Modality Earth Mover's Distance (cmEMD) to measure distance of two graphs of each modality and align these two modalities.

\textbf{Image generation based methods} commonly attempt to generate fake/virtual images that are used to bridge the modality gap in the image-level~\cite{wang2019learning,wang2019rgb,li2020infrared,choi2020hi,wang2020cross}. 
Wang et al.~\cite{wang2019learning} design a modality translation network to transfer the styles of images from one modality to another one. In addition, they unify the image and its transferred counterpart into a multi-spectral image to further reduce the appearance
discrepancy during representation learning. Similarly, Wang et al.~\cite{wang2019rgb} propose the Alignment Generative Adversarial Network (AlignGAN) to decrease intra-class variations in both pixel-level and feature-level. Instead of transferring images between two modalities, Li et al.~\cite{li2020infrared} produce an auxiliary X modality and convert the cross-modality learning as an X-Infrared-Visible three-mode learning problem. To reduce the impact of id-unrelated factors, several works (~\cite{choi2020hi,wang2020cross}) utilize the variational autoencoder and generative adversarial network to disentangle feature into id-related feature and id-unrelated feature, where the id-related feature is used for robust cross-modality matching. Despite the effectiveness of the above methods, they commonly require to train generative adversarial networks, which are difficult to optimize and sensitive to parameter settings.

\noindent\textbf{Earth Mover’s Distance (EMD)}~\cite{rubner1998emdmetric} is a metric to estimate the distance between two distributions, which is a special case of the transportation problem from linear optimization. EMD was applied to address many tasks, such as image retrieval~\cite{rubner2000earth}, tracking~\cite{zhao2008differential,li2013tensor,schulter2017deep}, graph matching~\cite{nikolentzos2017matching}, document retrieval~\cite{chen2019improving}, overcoming mode collapsing~\cite{chen2018adversarial,salimans2018improving,arjovsky2017wasserstein} of Generative Adversarial Network (GAN)~\cite{goodfellow2014generative}, and few-shot learning~\cite{zhang2020deepemd}. Different from 
them, this work is the first of employing EMD to solve the problem of VT-ReID, where we propose the cross-modality EMD to effectively learn modality-invariant representation.

\begin{figure*}[!t]
\centering
  \includegraphics[width=0.9\linewidth]{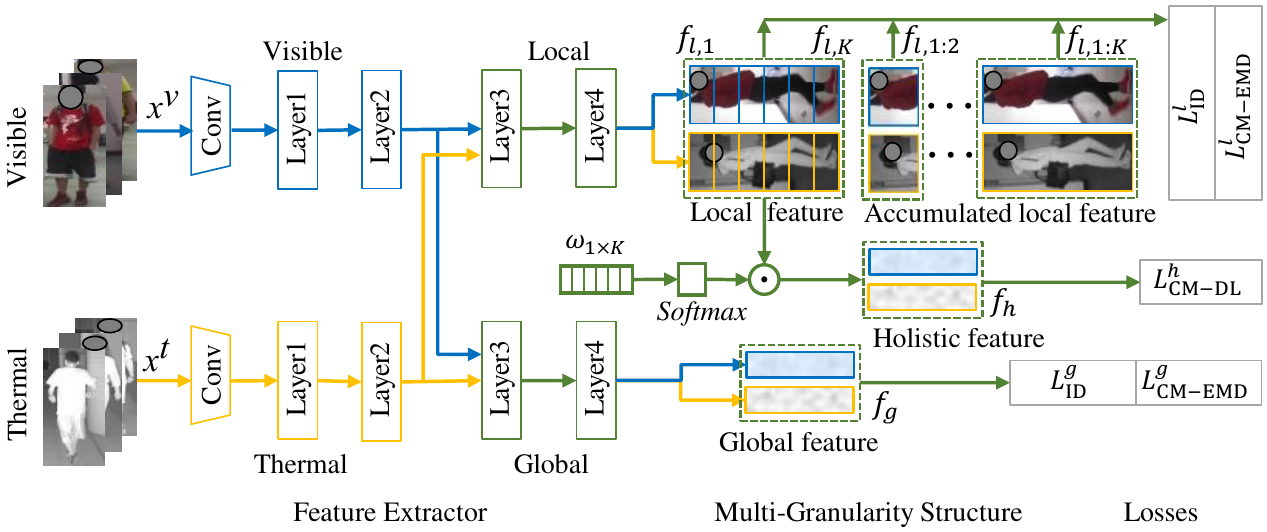}
  \caption{The framework of the proposed method. In the shallow layers, we use different parameters for visible modality and thermal modality. The shallow layers are the first convolutional layer, 1-th residual block and 2-th residual block of the ResNet-50~\cite{he2016deep}. After the shallow layers, we design two streams for extracting global-based features and local-based features. Each stream includes the 3-th and 4-th residual blocks of the ResNet-50~\cite{he2016deep}, which are shared by both modalities. The parameters of the two streams are different. Given the inputs, we first extract global feature, local features and accumulated local features with our Multi-Granularity Structure (MGS). We then calculate the losses of identity classification ($L_\mathrm{ID}^g$ and $L_\mathrm{ID}^l$) and the losses of our Cross-Modality Earth Mover’s Distance ($L_\mathrm{CM-EMD}^g$ and $L_\mathrm{CM-EMD}^l$) with the obtained three types of features. On the other hand, we generate the holistic feature by concatenating the weighted local features. The loss of our  Cross-Modality Discrimination Learning ($L_\mathrm{CM-DL}^h$) is calculated with the holistic feature.}
  \label{fig framework}
\end{figure*}

\section{Methodology}
\label{section3}

\textbf{Overview}. The framework of the proposed method is shown in Fig.~\ref{fig framework}. We first extract global feature, local features and accumulated local features with the proposed multi-granularity structure (MGS) for both visible and thermal modalities. We then calculate the losses of identity classification and losses of the proposed Cross-Modality Earth Mover’s Distance (CM-EMD) with the three types of features. For the proposed Cross-Modality Discrimination Learning (CM-DL), we first obtain a holistic feature by concatenating the weighted local features and compute the loss of CM-DL with the holistic feature. The loss of identity classification aims to learn a basic representation. The loss of CM-EMD is designed to effectively reduce the modality discrepancy. The loss of CM-DL focuses on learning more discriminative representation.

\subsection{Basic Loss}
\label{sec:basic}

Identity classification loss is widely used in ReID, which is obtained by calculating the cross-entropy loss with the identity labels. The identity classification loss is referred to $L_{ID}$ in this paper. Because the data in VT-ReID come from both visible and thermal modalities, we need to reduce the modality discrepancy during training so that the model can generate discriminative cross-modality representation. Next, we introduce a novel distribution alignment method to achieve the above goal.

\subsection{Cross-Modality Earth Mover's Distance}

\textbf{Motivation}. Distribution alignment is an effective way to reduce the modality discrepancy for VT-ReID. However, as discussed in the introduction, distribution alignment will be affected by the intra-identity variations. In our intuition, if we can select cross-modality pairs that have less intra-identity variations, the modality alignment process can mainly focus on reducing the modality discrepancies caused by the selected pairs. However, the above situation is hard to achieve since the intra-identity variations caused by various factors while we only have the identity information. On the other hand, EMD~\cite{rubner1998emdmetric} is a measure of the distance between two distributions, which can be solved by minimizing the cost of transporting one distribution to another. As a result, the two samples that are more similar will have a high connected weight otherwise have a low connected weight. Inspired by this, we introduce a new modality alignment method, named Cross-Modality Earth Mover's Distance (CM-EMD). Taking the advantage of EMD, CM-EMD can automatically assign weights between samples according to their similarities. As a consequence, the cost of CM-EMD is largely dominated by the pairs that have less intra-identity variations (\textit{i.e.}, have high similarities) and the impact of the intra-identity variations can be largely suppressed. We next introduce CM-EMD in detail.

Given a training mini-batch, we have $N^v$ visible modality samples and $N^t$ thermal modality samples, which are randomly selected from $C$ identities. The features obtained by the model are defined as $\mathbb{F}^v$ and $\mathbb{F}^t$ for visible modality and thermal modality, respectively. We will introduce how to obtain the representation of samples in Sec.~\ref{sec:mgs}.
The feature distributions of visible modality and thermal modality are denoted as $\boldsymbol{\nu} \in \mathbf{P}(\mathbb{F}^v)$ and $\boldsymbol{\tau} \in \mathbf{P}(\mathbb{F}^t)$, respectively. $\Pi(\boldsymbol{\nu}, \boldsymbol{\tau})$ represents all joint distributions $\rho(\boldsymbol{f^v}, \boldsymbol{f^t})$, where $\boldsymbol{f^v} \in \mathbb{F}^v$  and $\boldsymbol{f^t} \in \mathbb{F}^t$. The CM-EMD among the samples of two modalities can be defined as:
\begin{equation}
\small
\label{eq:emd}
\mathcal{D}_\mathrm{CM-EMD}(\mathbb{F}^v, \mathbb{F}^t)=\inf_{\boldsymbol{\rho} \in \Pi(\boldsymbol{\nu}, \boldsymbol{\tau})} \mathbb{E}_{(\boldsymbol{f^v}, \boldsymbol{f^t}) \sim \boldsymbol{\rho}}[M(\boldsymbol{f^v}, \boldsymbol{f^t})],
\end{equation}
\noindent where $M(\boldsymbol{f^v}, \boldsymbol{f^t})$ is the cost function, which is calculated by the euclidean distance between $\boldsymbol{f^v}$ and $\boldsymbol{f^t}$. EMD has the form of transportation problem (TP) from Linear Programming, and Eq.~\ref{eq:emd} can be re-formulated as:
\vspace{-.05in}
\begin{equation}
\small
\label{eq:emd-tp}
\mathcal{D}_\mathrm{CM-EMD}(\mathbb{F}^v, \mathbb{F}^t)=\min _{\mathbf{S} \in \Pi(\mathbf{V}, \mathbf{T})} \sum_{i=1}^{N^v} \sum_{j=1}^{N^t} \mathbf{S}_{i j} \cdot M\left(\boldsymbol{f^v_{i}}, \boldsymbol{f^t_{j}}\right),
\end{equation}
where $\mathbf{V}=\{{v}_{i}\}_{i=1}^{N^v}$ and $\mathbf{T}=\{{t}_{i}\}_{i=1}^{N^t}$ are $N^v$-dim simplex and $N^t$-dim simplex, respectively. $v_{i}$ and $t_{j}$ are the weights of their corresponding nodes. $\Pi(\mathbf{V}, \mathbf{T})$ denotes all transport plans $\mathbf{S} \in \mathbb{R}_{+}^{N^v \times N^t}$. $S_{i j}$ indicates the weight of shifting from $v_{i}$ to $t_{j}$, which subjects to:
\vspace{-.05in}
\begin{equation}
\small
\label{eq:emd-subject}
\begin{aligned}
\quad S_{i j} \geqslant 0, ~~~~\sum_{j=1}^{N^t} S_{i j}=v_{i}, ~~~~\sum_{i=1}^{N^v} S_{i j}=t_{j}, \\
\forall i=1, \ldots, N^v, ~~~~\forall j=1, \ldots, N^t.
\end{aligned}
\end{equation}
To the reduce computational complexity, we use Sinkhorn algorithm~\cite{cuturi2013sinkhorn, peyre2019computational} to solve the Eq.~\ref{eq:emd-tp}.

\begin{figure}[!t]
\centering
  \includegraphics[width=0.98\linewidth]{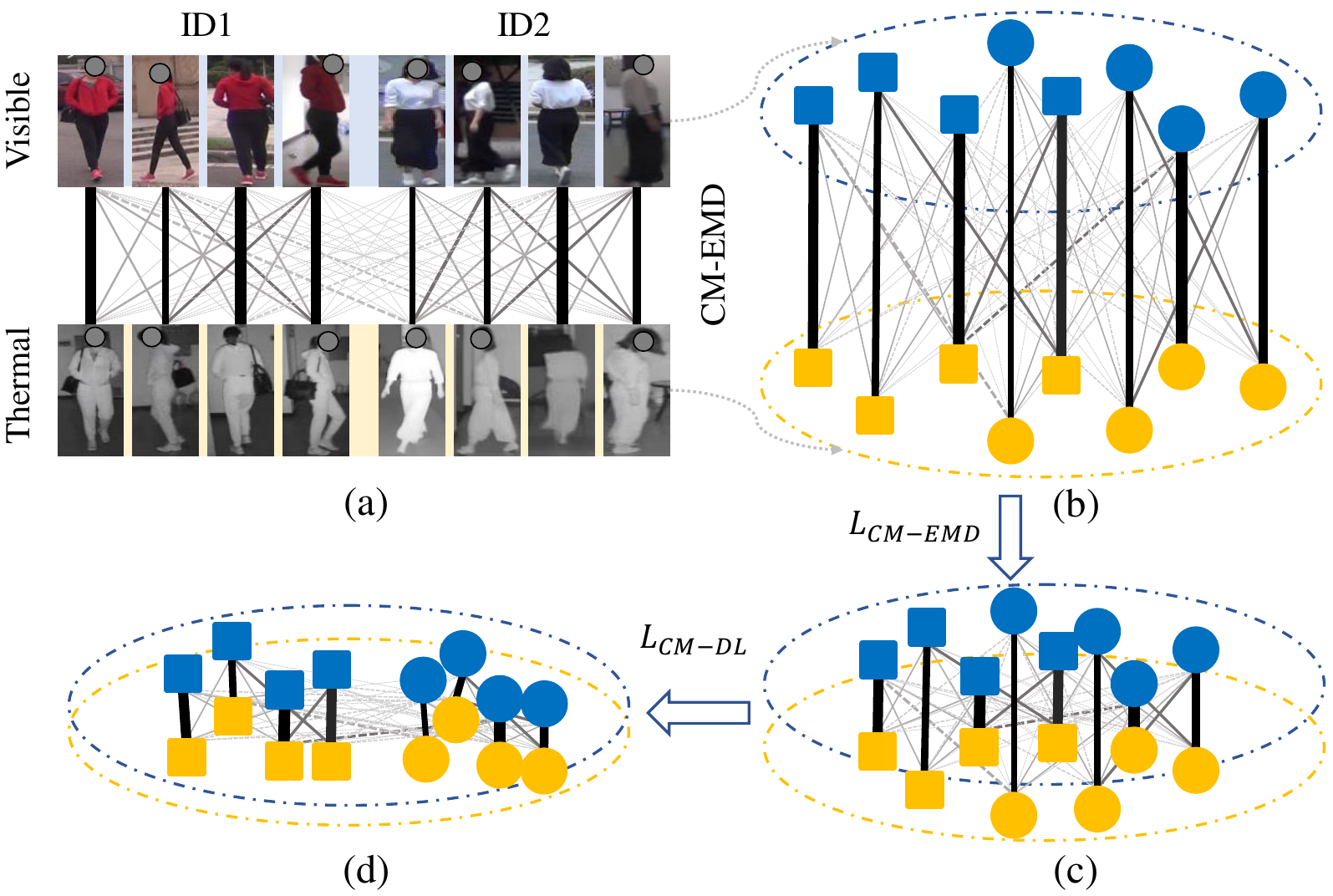}
  \caption{
  Illustration of the proposed CM-EMD (a$\rightarrow$c) and CM-DL (c$\rightarrow$d). CM-EMD can effectively reduce the modality gap while CM-DL can effectively promote the discrimination ability of the model. The pairs with high weights are connected with wider lines. Shapes indicate the identities. Colors represent the modalities (blue for visible and yellow for thermal).
  }
  \label{fig emd}
\end{figure}

\textbf{Discussion}. After optimization, there is a negative correlation between the transport weight $S_{i j}$ and the transport cost $M(\boldsymbol{f^v_{i}}, \boldsymbol{f^t_{j}})$. That is, the two samples, $f^v_{i}$ and $f^t_{j}$, will be assigned with a large transport weight when they are close to each other, otherwise will be assigned with a small transport weight. As a result, when using Eq.~\ref{eq:emd-tp} as the loss function of modality alignment, the model will mainly focus on reducing the discrepancy of samples that have less intra-identity variations (\textit{i.e.}, have small transport cost). Meanwhile, the samples that have large intra-identity variations will have less impact on the model optimization, since they are assigned with low weights. An example of CM-EMD is illustrated in Fig.~\ref{fig emd} (a$\rightarrow$c).

\subsection{Cross-Modality Discrimination Learning}

One disadvantage of modality alignment is that the optimization process may degrade the discrimination capability of the model, which is also a problem of the proposed CM-EMD. Class-wise variance can well represent the identity distributions. Specifically, if the model can well distinguish samples of different identities, the intra-class variance should be small while the inter-class variance should be large. Taking this into mind, we propose the  Cross-Modality Discrimination Learning (CM-DL) to improve the discrimination ability of the model, which is achieved by constraining the relation between cross-modality intra-class variance and cross-modality inter-class variance. 

Given a training mini-batch, we first calculate the feature mean for each modality, formulated as:
\begin{equation}\label{eq mean-modality}
\small
\mu^{v}=\frac{1}{N^{v}} \sum_{i=1}^{N^{v}} f_{i}^{v}, \quad \mu^{t}=\frac{1}{N^{t}} \sum_{i=1}^{N^{t}} f_{i}^{t}.
\end{equation}
We then calculate the class-wise feature mean for each modality, formulated as:
\begin{equation}\label{eq mean-class}
\small
\mu_{c}^{v}=\frac{1}{N_{c}^{v}} \sum_{i=1}^{N_{c}^{v}} f_{i, c}^{v}, \quad \mu_{c}^{t}=\frac{1}{N_{c}^{t}} \sum_{i=1}^{N_{c}^{t}} f_{i, c}^{t},
\end{equation}
where ${N_{c}^{v}}$ and ${N_{c}^{t}}$ denote the number of samples of class $c$ for visible modality and thermal modality, respectively. $f_{i, c}$ is the feature of the sample that belongs to the class $c$.

The cross-modality intra-class variance is defined as:
\begin{equation}\label{eq vw}
\small
\begin{aligned}
V_{intra}(\mathbb{F}^v, \mathbb{F}^t)=\sum_{c=1}^{C}&\left\{ \sum_{i=1}^{N_{c}^{t}}\left(f_{i, c}^{t}-\mu_{c}^{v}\right)\left(f_{i, c}^{t}-\mu_{c}^{v}\right)^\top\right. \\
&\left.+\sum_{i=1}^{N_{c}^{v}}\left(f_{i, c}^{v}-\mu_{c}^{t}\right)\left(f_{i, c}^{v}-\mu_{c}^{t}\right)^\top\right\},
\end{aligned}
\end{equation}
where $^\top$ represents the transpose operation.
Similarly, the cross-modality inter-class variance is defined as:
\begin{equation}\label{eq vb}
\small
\begin{aligned}
V_{inter}(\mathbb{F}^v, \mathbb{F}^t)=\sum_{c=1}^{C}&\left\{N_{c}^{v}\left(\mu_{c}^{v}-\mu^{t}\right)\left(\mu_{c}^{v}-\mu^{t}\right)^\top\right. \\
&\left.+N_{c}^{t}\left(\mu_{c}^{t}-\mu^{v}\right)\left(\mu_{c}^{t}-\mu^{v}\right)^\top\right\}.
\end{aligned}
\end{equation}

By jointly considering the cross-modality intra-class variance and cross-modality inter-class variance, our CM-DL can be formulated as:
\begin{equation}\label{eq CM-DL}
\small
    L_\mathrm{CM-DL}=\frac{V_{intra}(\mathbb{F}^v, \mathbb{F}^t)}{V_{inter}(\mathbb{F}^v, \mathbb{F}^t)}.
\end{equation}

By minimizing Eq.~\ref{eq CM-DL}, the model is encouraged to reduce the cross-modality intra-class variance as well as increase the cross-modality inter-class variance, leading to a more discriminative representation. An example of CM-DL is shown in Fig.~\ref{fig emd} (c$\rightarrow$d).

\subsection{Multi-Granularity Structure}
\label{sec:mgs}

In the community of ReID~\cite{sun2018beyond}, it is has been demonstrated that local information is critical for extracting discriminative pedestrian representation. Inspired by this, we propose the multi-granularity structure (MGS) to obtain more discriminative feature as well as align the modalities in both coarse- and fine-grained aspects.

Given the feature of the last residual convolution layer $f_{res} \in \mathbb{R}^{H, W, C}$, we extract three types of features: global feature, local feature, accumulated local feature, which are introduced in the followings. Note that, since we adopt different streams to obtain $f_{res}$ for global feature and local-based features, we use $f_{g, res}$ for global stream and $f_{l, res}$ for local stream.

\textbf{Global Feature}. We directly apply the generalized-mean pooling (GeM)~\cite{radenovic2018fine} and the batch normalization neck (BNNeck)~\cite{luo2019strong} on $f_{g, res}$ and obtain the global feature $f_{g} \in \mathbb{R}^{1, C}$.

\textbf{Local Feature}. We first evenly divide $f_{l, res}$ into $K$ parts along the vertical direction and perform the GeM and BNNeck on them, obtaining $K$ local features, $\{f_{l, 1}, ..., f_{l, K}\}$.

\textbf{Accumulated Local Feature}. To obtain more diverse features, we extract accumulated local features by combining different number of part features, which are represented by $\{f_{l, 1:2}, ..., f_{l, 1:K}\}$. For example, $f_{l, 1:2}$ is obtained by concatenating the local features $f_{l, 1}$ and $f_{l, 2}$.

\textbf{Losses for CM-EMD}. Given the multi-granularity features, the losses of CM-EMD can be calculated as:
\begin{equation}\label{eq loss-emd}
\small
\begin{aligned}
L_\mathrm{CM-EMD}^{g} &=D_\mathrm{CM-EMD}\left(\mathbb{F}_{g}^{v}, \mathbb{F}_{g}^{t}\right), \\
L_\mathrm{CM-EMD}^{l} &=\sum_{k=1}^{K} D_\mathrm{CM-EMD}\left(\mathbb{F}_{l,k}^{v}, \mathbb{F}_{l,k}^{t}\right) \\
&+\alpha \sum_{k=2}^{K} D_\mathrm{CM-EMD}\left(\mathbb{F}_{l, 1:k}^{v}, \mathbb{F}_{l, 1:k}^{t}\right),
\end{aligned}
\end{equation}
where $\alpha$ is the hyper-parameter and $K$ is the part number. $L_\mathrm{CM-EMD}^{g}$ is the global-based loss of CM-EMD and $L_\mathrm{CM-EMD}^{l}$ is the local-based loss of CM-EMD.

\textbf{Loss for CM-DL}. Instead of using the global-based and local-based features individually, we concatenate the weighted local features to produce a holistic feature, which is utilized to the calculated loss of CM-DL. Specifically, the holistic feature is represented as:
\begin{equation}\label{eq fh}
\small
f_{h}= [\omega_{1}f_{l, 1}\mid\omega_{2} f_{l,2}\mid \cdots \mid \omega_{K} f_{l, K}],
\end{equation}
where $[\cdot\mid\cdot]$ indicates the concatenation operation. $\omega_1, ..., \omega_{K}$ are the trainable weights, which are normalized by the SoftMax function.

The loss of CM-DL is re-formulated as follows:
\begin{equation}\label{eq CM-DL-h}
\small
    L_\mathrm{CM-DL}^h=\frac{V_{intra}(\mathbb{F}_h^v, \mathbb{F}_h^t)}{V_{inter}(\mathbb{F}_h^v, \mathbb{F}_h^t)}.
\end{equation}

\textbf{Losses for Identity Classification}. As explained in Sec.~\ref{sec:basic}, $L_{ID}$ is calculated by identity classification loss. We apply $L_{ID}$ on the global feature, local features, and accumulated local features. Specifically, we add classification heads (1 for global feature, $K$ for local features, and $K$ for accumulated local features), which are fully-connected layers with output length of $\#$identities, after these three types of features. We then calculated identity classification losses according to the outputs of these classification heads, formulated as:
\begin{equation}\label{eq loss-ID}
\small
\begin{aligned}
L_\mathrm{ID}^{g} &=\mathrm{CrossEntropy}\left(\mathbb{H}_{g}, \mathbb{Y}\right), \\
L_\mathrm{ID}^{l} &=\sum_{k=1}^{K} \mathrm{CrossEntropy}\left(\mathbb{H}_{l,k}, \mathbb{Y}\right) \\
&+\alpha \sum_{k=2}^{K} \mathrm{CrossEntropy}\left(\mathbb{H}_{l, 1:k}, \mathbb{Y}\right),
\end{aligned}
\end{equation}
where $\mathbb{H}$ indicates the predictions of the corresponding classification heads and $\mathbb{Y}$ is the identity labels.

\subsection{Overall}

\textbf{Training}. By considering the proposed CM-EMD, CM-DL, MSG and the basic loss, the model is optimized by:
\begin{equation}\label{eq overall}
\small
\begin{aligned}
  \arg \min_{\theta}  \gamma_1 L_\mathrm{CM-DL}^h + \gamma_2 L_\mathrm{ID}^l + \gamma_3 L_\mathrm{CM-EMD}^l \\  + \gamma_4 L_\mathrm{ID}^g + \gamma_5 L_\mathrm{CM-EMD}^g,
\end{aligned}
\end{equation}
where $\gamma_{1 \to 5}$ are the hyper-parameters that balance the importance of different losses. 

\textbf{Testing}. In the testing phase, we obtain the final feature by concatenating the global feature and local features:
\begin{equation}\label{eq test}
\small
\begin{aligned}
f_{test}&=[\beta f_{l,1:K} \mid (1-\beta) f_g],
\end{aligned}
\end{equation}
where $\beta$ is the hyper-parameter that controls the importance of global and local features.

\begin{table*}[!t]
  \centering
  \caption{Comparison with the state-of-the-art methods on the RegDB dataset.} 
    \begin{tabular}{l|cccc|cccc}
    \hline
    Settings &
    \multicolumn{4}{c|}{Visible to Thermal}  & \multicolumn{4}{c}{Thermal to Visible} \\
    \hline
    Method & R1    & R10   & R20   & mAP  & R1    & R10   & R20   & mAP \\
    \hline
    HOG~\cite{dalal:inria-00548512}   & 13.49 & 33.22 & 43.66 & 10.31 & / 	& / 	& / 	& / \\
    LOMO~\cite{liao2015person}  & 0.85  & 2.47  & 4.1   & 2.28 & / 	& / 	& / 	& / \\
    One-stream~\cite{wu2017rgb}  & 13.11 & 32.98 & 42.51 & 14.02 & / 	& / 	& / 	& /  \\
    Two-stream~\cite{wu2017rgb}  & 12.43 & 30.36 & 40.96 & 13.42 & / 	& / 	& / 	& /  \\
    Zero-Padding~\cite{wu2017rgb} & 17.75 & 34.21 & 44.35 & 18.9 & / 	& / 	& / 	& / \\
    HCML~\cite{ye2018hierarchical} &24.44 & 47.53 & 56.78 & 20.8 & 21.70 & 45.02 & 55.58 & 22.24 \\
    BDTR~\cite{ye2018visible} & 33.47 & 58.42 & 67.52 & 31.83 & 32.92 & 58.46 & 68.43 & 31.96 \\

    DGD+MSR~\cite{feng2019learning} & 48.43 & 70.32 & 79.95 & 48.67 & / 	& / 	& / 	& / \\
    D2RL~\cite{wang2019learning}  & / 	& / 	& / 	& / & 43.4  & 66.1  & 76.3  & 44.1 \\
    EDFL~\cite{liu2020enhancing} & 52.58 & 72.1  & 81.47 & 52.98 & 51.89 & 72.09 & 81.04 & 52.13 \\
    D-HSME~\cite{hao2019hsme} & 50.85 & 73.36 & 81.66 & 47 & 50.15 & 72.40 & 81.07 & 46.16  \\
    AlignGAN~\cite{wang2019rgb} & 57.9 & /     & / &53.6 & 56.3  & /     & /     & 53.4 \\
    DFE~\cite{hao2019dual} 	& 70.13	& 86.32	& 91.96	& 69.14	& 67.99	& 85.56	&  91.41	&  66.70 \\
    Hi-CMD~\cite{choi2020hi} & / 	& / 	& / 	& / & 70.93	& 86.39	& /	& 66.04		 \\
    PIG~\cite{wang2020cross} & 48.50	& /	& /	& 49.3	 & 48.1	& /	& /	& 48.90	\\
    Xmodal~\cite{li2020infrared} & 62.21 & 83.13	& 91.72	& 60.18	& / 	& / 	& / 	& /   \\
    CMM-CML~\cite{ling2020class}  & 59.81 & 80.39 & 88.69 & 60.86 & / & / & / & / \\
    DDAG~\cite{ye2020dynamic} 	& 69.34 	& 86.19 	& 91.49 	& 63.46 & 68.06	& 85.15	& 90.31	& 61.80 \\
    DG-VAE~\cite{pu2020dual} & 72.97 & / & 86.89 & 71.78	& /		& / 	& / 	& / \\ 
    cm-SSFT~\cite{lu2020cross}	& 73.3	& /	& /	& 72.9	& 71.0	& /	& /	& 71.7 \\
    SIM~\cite{jia2020similarity} & 75.29	& /	& /	& 74.47	& 78.30 	& / 	& / 	& 75.24 \\

    AGW~\cite{ye2021deep} & 70.05  & 86.21   & 91.55   & 66.37  & 70.49  & 87.12     & 91.84  & 65.90 \\
    CMAlign~\cite{park2021learning} & 74.17	& /	& /	& 67.64	& 72.43 	& / 	& / 	& 65.46 \\
    SFANet~\cite{liu2021sfanet} & 76.31 & 91.02 & 94.27 & 68.00  & 70.15 & 85.24 & 89.27 & 63.77 \\
    NFS~\cite{chen2021neural} & 80.54 & 91.96 & 95.07 & 72.1 & 77.95 & 90.45 & 93.62 & 69.79 \\
    VCD-VML~\cite{tian2021farewell}  & 73.2  & /   & /   & 71.6  & 71.8  & /     & /  & 70.1 \\
    MCLNet~\cite{hao2021cross} & 80.3  & 92.7  & 96.03 & 73.07 & 75.93 & 90.93 & 94.59 & 69.49 \\
    MPANet~\cite{wu2021discover}  & 83.7  & /     & /     & 80.9 & 82.8  & /     & /     & 80.7 \\
    SMCL~\cite{wei2021syncretic} & 83.93	& /	& /	& 79.83	& 83.05 	& / 	& / 	& 78.57 \\
    CM-NAS~\cite{fu2021cm} & 84.54 & 95.18 & 97.85 & 80.32  & 82.57 & 94.51 & 97.37 & 78.31 \\
    \hline
    \textbf{Ours}  & \textbf{94.37} & \textbf{98.93} & \textbf{99.42} & \textbf{88.23} & \textbf{92.77} & \textbf{98.50} & \textbf{99.66} & \textbf{86.85} \\

    \hline
    \end{tabular}%
    \vspace{-.1in}
  \label{tab sota-reg}%
\end{table*}%

\section{Experiment}
\label{section4}

\subsection{Experimental settings}
\textbf{Datasets.} 
Experiments are conducted on two VT-ReID datasets, \textit{i.e.}, SYSU-M001~\cite{wu2017rgb} and RegDB~\cite{nguyen2017person}.

\textit{SYSU-M001}~\cite{wu2017rgb} contains 287,628 RGB images and 15,729 infrared images, which are captured by four RGB cameras and two thermal cameras, respectively. The training set contains 22,258 RGB images and 11,909 infrared images of 395 identities. The testing set involves 3,803 query (infrared) images and 301 gallery (RGB) images of 96 identities. For evaluation, we use two testing modes, \textit{i.e.,}, all-search mode and indoor-search mode, we report single-shot setting for these two modes.

\textit{RegDB}~\cite{nguyen2017person} comprises 4,120 RGB images and 4,120 infrared images of 412 identities, collected from one RGB camera and one infrared camera. Each identity has 10 RGB images and 10 infrared images. For evaluation, we equally divide RegDB into the training  and testing sets. Two testing settings are used: Thermal (query) to Visible (gallery) setting and Visible (query) to Thermal (gallery) setting.

\textbf{Evaluation metrics}. The Cumulative Matching Characteristics (CMC) and mean Average Precision (mAP) are used to evaluate the retrieval performance. For CMC, we report the rank-1 (R1), rank-10 (R10), and rank-20 (R20) accuracies.

\textbf{Implementation Details}. The baseline model is trained with only the identity classification loss. We use random cropping for data augmentation during training. For each training mini-batch, we set the number of identities to 6 (\textit{i.e.}, $C$=6) for both datasets. We then randomly sample 8 RGB images and 8 infrared images (\textit{i.e.}, $N_c^v=N_c^t=8$) for SYSU-M001, and 4 RGB images and 4 infrared images (\textit{i.e.}, The input images are resized to $384 \times 192 \times3$. We choose the ResNet-50~\cite{he2016deep} as the backbone of the feature extractor. The SGD optimizer is used to update the parameters of the network, where we set initial learning rate to 0.01 and divide the learning rate by 10 after every 30 epochs. We train the model for a total of 80 epochs. For the hyper-parameters of the proposed model, we set $\alpha$ (Eq.~\ref{eq loss-emd}) to 0.2 and 1.0 for SYSU-M001 and RegDB, respectively. $\gamma_{1\to5}$ (Eq.~\ref{eq overall}) are to \{1, 1, 0.1, 2, 0.1\} for SYSU-M001 and \{3, 2, 0.4, 1, 0.6\} for RegDB, respectively. During testing, we set $\beta$ (Eq.~\ref{eq test}) to 0.7 and 0.5 for SYSU-M001 and RegDB, respectively.

\subsection{Comparison with The State of The Art}

To demonstrate the superiority of our method, we compare it against the state-of-the-art approaches on SYSU-MM01 and RegDB. The competitors include feature extraction based methods (Zero-Padding~\cite{wu2017rgb},BDTR~\cite{ye2018visible},cm-SSFT~\cite{lu2020cross}, SIM~\cite{jia2020similarity}, DDAG~\cite{ye2020dynamic}, NFS~\cite{chen2021neural}, VCD-VML~\cite{tian2021farewell}), metric learning based methods (HCML~\cite{ye2018hierarchical},D-HSME~\cite{hao2019hsme}, BFE+HPI~\cite{zhao2019hpiln}, DGD+MSR~\cite{feng2019learning}, CMM-CML~\cite{ling2020class}, MCLNet~\cite{hao2021cross}), distribution alignment based methods (cmGAN~\cite{dai2018cross},DFE~\cite{hao2019dual}, DG-VAE~\cite{pu2020dual}, MPANet~\cite{wu2021discover}), and image generation based methods (D2RL~\cite{wang2019learning}, AlignGAN~\cite{wang2019rgb}, Xmodal~\cite{li2020infrared}, Hi-CMD~\cite{choi2020hi}, PIG~\cite{wang2020cross}).

%

\begin{table*}[!ht]
  \centering
  \caption{Comparison with the state-of-the-art methods on the SYSU-MM01 dataset.}
    \begin{tabular}{l|llll|llll}
    \hline
    \multicolumn{1}{l|}{Settings}
    & \multicolumn{4}{c|}{All-search }  & \multicolumn{4}{c}{Indoor-search} \\
    \hline
    Method & R1    & R10   & R20   & mAP   & R1    & R10   & R20   & mAP  \\
    \hline
    HOG~\cite{dalal:inria-00548512}   & 2.76  & 18.25 & 31.91 & 4.24   & 3.22  & 24.68 & 44.52 & 7.25  \\
    LOMO~\cite{liao2015person}  & 3.64  & 23.18 & 37.28 & 4.53  & 5.75  & 34.35 & 54.9  & 10.19  \\
    One-stream~\cite{wu2017rgb} & 12.04 & 49.68 & 66.74 & 13.67   & 16.94 & 63.55 & 82.1  & 22.95 \\
    Two-stream~\cite{wu2017rgb} & 11.65 & 47.99 & 65.5  & 12.85  & 15.6  & 61.18 & 81.02 & 21.49 \\
    Zero-padding~\cite{wu2017rgb} & 14.8  & 54.12 & 71.33 & 15.95 & 20.58 & 68.38 & 85.79 & 26.92  \\
    HCML~\cite{ye2018hierarchical}  & 14.32 & 53.16 & 69.17 & 16.16      & /     & /     & /     & / \\
    cmGAN~\cite{dai2018cross} & 26.97 & 67.51 & 80.56 & 27.8   & 31.63 & 77.23 & 89.18 & 42.19  \\
    BDTR~\cite{ye2018visible}  & 17.01 & 55.43 & 71.96 & 19.66     & /     & /     & /     & / \\
    D-HSME~\cite{hao2019hsme} & 20.68 & 62.74 & 77.95 & 23.12      & /     & /     & /     & / \\
    D2RL~\cite{wang2019learning}  & 28.9  & 70.6  & 82.4  & 29.2   & /     & /     & /     & / \\
    DGD+MSR~\cite{feng2019learning} & 37.35 & 83.4  & 93.44 & 38.11  & 39.64 & 89.29 & 97.66 & 50.88  \\
    BFE+HPI~\cite{zhao2019hpiln} & 41.36 & 84.78 & 94.51 & 42.95 & 45.77 & 91.82 & 98.46 & 56.52 \\
    AlignGAN~\cite{wang2019rgb} & 42.4  & 85    & 93.7  & 40.7   & 45.9  & 87.6  & 94.4  & 54.3  \\
    DFE~\cite{hao2019dual} & 48.71	& 88.86	& 95.27	& 48.59	& 52.25 & 89.86	& 95.85	& 59.68	\\
    EDFL~\cite{liu2020enhancing}  & 36.94 & 84.52 & 93.22 & 40.77   & /     & /     & /     & / \\
    Hi-CMD~\cite{choi2020hi} & 34.94	& 77.58	& /	& 35.94		& / 	& / 	& / 	& / \\
    PIG~\cite{wang2020cross} & 38.1	& 80.7	& 89.9	& 36.9	& 43.8	& 86.2	& 94.2	& 52.9	\\
    Xmodal~\cite{li2020infrared} & 49.92 & 89.79	& 95.96	& 50.73	 	& / 	& / 	& / 	& / \\
    CMM-CML~\cite{ling2020class}  & 51.8 & 92.72 & 97.71 & 51.21 & 54.98 & 94.38 & 99.41 & 63.7  \\
    DDAG~\cite{ye2020dynamic} & 54.75	& 90.39	& 95.81	& 53.02	& 61.02	& 94.06	& 98.41	& 67.98	\\
    DG-VAE~\cite{pu2020dual}	& 59.49	& /	& 93.77	& 58.46		& / 	& / 	& / 	& / \\
    SIM~\cite{jia2020similarity} & 60.88	& /	& /	& 56.93	& / 	& / 	& / 	& / \\
    cm-SSFT~\cite{lu2020cross}	& 61.6	& 89.2	& 93.9	& 63.2		& 70.5	& 94.9	& 97.7	& 72.6	 \\
    AGW~\cite{ye2021deep} & 47.50 & 84.39	& 92.14	& 47.65	& 54.17 	& 91.14 	& 95.98 	& 62.97  \\
    CMAlign~\cite{park2021learning} & 55.41	& /	& /	& 54.14	& 58.46 	& / 	& / 	& 66.33 \\
    NFS~\cite{chen2021neural}	& 56.91 & 91.34 & 96.52 & 55.45  & 62.79 & 96.53 & 99.07 & 69.79  \\
    VCD-VML~\cite{tian2021farewell}	& 60.02 & 94.18 & 98.14 & 58.8  & 66.05 & 96.59 & 99.38 & 72.98  \\
    CM-NAS~\cite{fu2021cm} & 61.99 & 92.87 & 97.25 & 60.02  & 67.01 & 97.02 & 99.32 & 72.95\\
    MC-AWL~\cite{lingmulti} & 64.82	& /	& /	& 60.81	& / 	& / 	& / 	& / \\
    SFANet~\cite{liu2021sfanet} & 65.74 & 92.98 & 97.05 & 60.83 		& 71.60 & 96.60 & 99.45 & 80.05 \\
    MCLNet~\cite{hao2021cross} & 65.4  & 93.33 & 97.14 & 61.98 & 72.56 & 96.98 & 99.2  & 76.58  \\
    SMCL~\cite{wei2021syncretic} & 67.39 & 92.87 & 96.76 & 61.78 	& 68.84 & 96.55 & 98.77 & 75.56 \\
    MPANet~\cite{wu2021discover}	& 70.58 & 96.21 & 98.8  & 68.24 & 76.74 & 98.21 & 99.57 & 80.95  \\
    \hline
    \textbf{Ours}  & \textbf{73.39} & \textbf{96.24} & \textbf{98.82} & \textbf{68.56}  & \textbf{80.53} & \textbf{98.31} & \textbf{99.91} & \textbf{82.71}  \\
    \hline
    \end{tabular}
  \label{tab sota-sysu}%
\end{table*}%

\textbf{Results on RegDB}. We first compare our method with state-of-the-art methods on RegDB. The results are reported in Tabel~\ref{tab sota-reg}. It is clear that our method outperforms the state-of-the-art methods by a large margin on both evaluation settings. Specifically, we obtain \textbf{rank-1 accuracy = 94.37\%} and \textbf{mAP accuracy = 88.23\%} for the ``Visible to Thermal'' setting, and, \textbf{rank-1 accuracy = 92.77\%} and \textbf{mAP accuracy = 86.85\%} for ``Thermal to Visible'' setting. Compared to the current best competitor (CM-NAS~\cite{fu2021cm} and SMCL~\cite{wei2021syncretic} published in ICCV 2021), we show significant improvements. For example, our method is higher than CM-NAS by 9.83\% in rank-1 accuracy and 7.91\% in mAP accuracy on the ``Visible to Thermal'' setting, and, outperforms SMCL by 9.72\% in rank-1 accuracy and 8.28\% in mAP accuracy on the ``Thermal to Visible'' setting, respectively. This indicates that our method leads a new state-of-the-art performance on RegDB.

\textbf{Results on SYSU-MM001}. The comparisons on SYSU-MM001 are reported in  Tabel~\ref{tab sota-sysu}. In all settings, our method achieves the best results in rank-1 accuracy and obtains competitive results in mAP. Specifically, we achieve \textbf{rank-1 accuracy = 73.39\%} and \textbf{mAP accuracy = 68.56\%} for the all-search mode, and, \textbf{rank-1 accuracy = 80.53\%} and \textbf{mAP accuracy = 82.71\%} for the indoor-search mode. 
Compared to the current best competitor (MPANet~\cite{wu2021discover} published in CVPR 2021), our method clearly surpasses MPANet by 2.81\% and 3.79\% in rank-1 accuracy on the two evaluation settings, respectively.

\begin{table}[!t]
  \centering
  \setlength{\tabcolsep}{5pt}
  \small
  \caption{Ablation study of the proposed components. \textbf{CM-EMD}: cross-modality earth mover’s distance, \textbf{CM-DL}: Cross-Modality Discrimination Learning, \textbf{Global}: global-based feature, \textbf{Local}: local based features, \textbf{MGF}: global-based \& local-based features.}
    \begin{tabular}{l|l|cc|cc}
    \hline
    \multirow{2}[1]{*}{\#}&\multirow{2}[1]{*}{Method} & \multicolumn{2}{c|}{SYSU-M001} & \multicolumn{2}{c}{RegDB} \\
    \cline{3-6} 
    & & R1    & mAP   & R1    & mAP  \\
    \hline
    0&Baseline w/ Global &54.22  & 51.18  & 46.46  & 43.48  \\
    1&+CM-EMD& \bf 60.01  & \bf 54.75  & \bf 76.76  & \bf 69.70 \\
    \hline
    2&Baseline w/ Local   & 55.56  & 53.00  & 64.85  & 61.90 \\
    3&+CM-EMD& 65.97  & 62.34  & 89.37 & 81.19 \\
    4&+CM-EMD+CM-DL& \bf 67.81  & \bf 63.68  & \bf 93.45  & \bf 85.33 \\
    \hline
    5&Baseline w/ MGF & 61.58  & 59.26  & 69.08 & 64.22 \\
    6&+CM-EMD & 71.26  & 66.59  & 92.86 & 84.99 \\
    7&+CM-EMD+CM-DL & \bf 73.39  & \bf 68.56  & \bf 94.37  & \bf 88.23 \\
    \hline
    \end{tabular}%
  \label{tab:ablation}%
\end{table}%

\begin{table}[ht]
\vspace{-.1in}
\caption{Effect of accumulated local features (ALF).}
\centering
  \footnotesize
    \begin{tabular}{l|cc|cc}
    \hline
    \multirow{2}[1]{*}{Method} & \multicolumn{2}{c|}{RegDB} & \multicolumn{2}{c}{SYSU-M001} \\
\cline{2-5}          & R1    & mAP   & R1    & mAP  \\
    \hline
    Ours w/o ALF & 91.94  & 84.95  & 72.39  & 67.78  \\
    Ours & \bf 94.37  & \bf 88.23 & \bf 73.39  & \bf 68.56  \\
    \hline
    \end{tabular}%
  \label{tab: acc}%
  \vspace{-.15in}
\end{table}%

\begin{figure*}[!t]
\centering
  \includegraphics[width=0.95\linewidth]{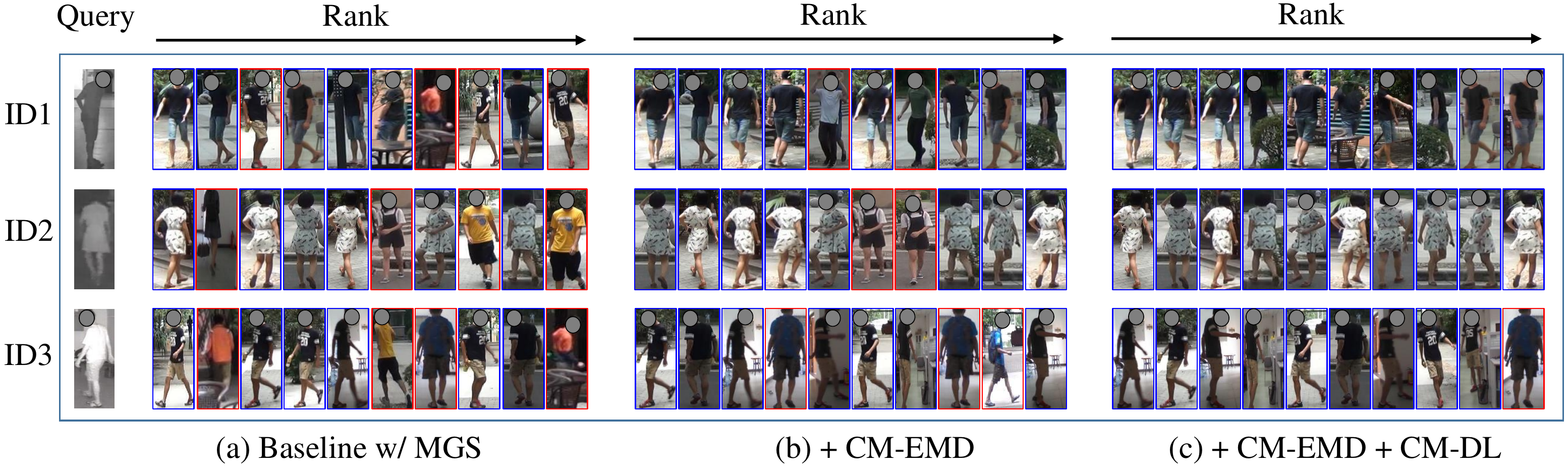}
    \caption{Comparison of the top-10 retrieval results of the (a) Baseline w/ MGS, (b) Baseline w/ MGS + CM-EMD and (c) Baseline w/ MGS + CM-EMD + CM-DL. We evaluate the models on the SYSU-M001 dataset. Sample with blue / red box is positive / negative (best view in zoom).}
  \label{fig retrieval}
\end{figure*}

\begin{figure*}[!t]
\centering
  \includegraphics[width=0.90\linewidth]{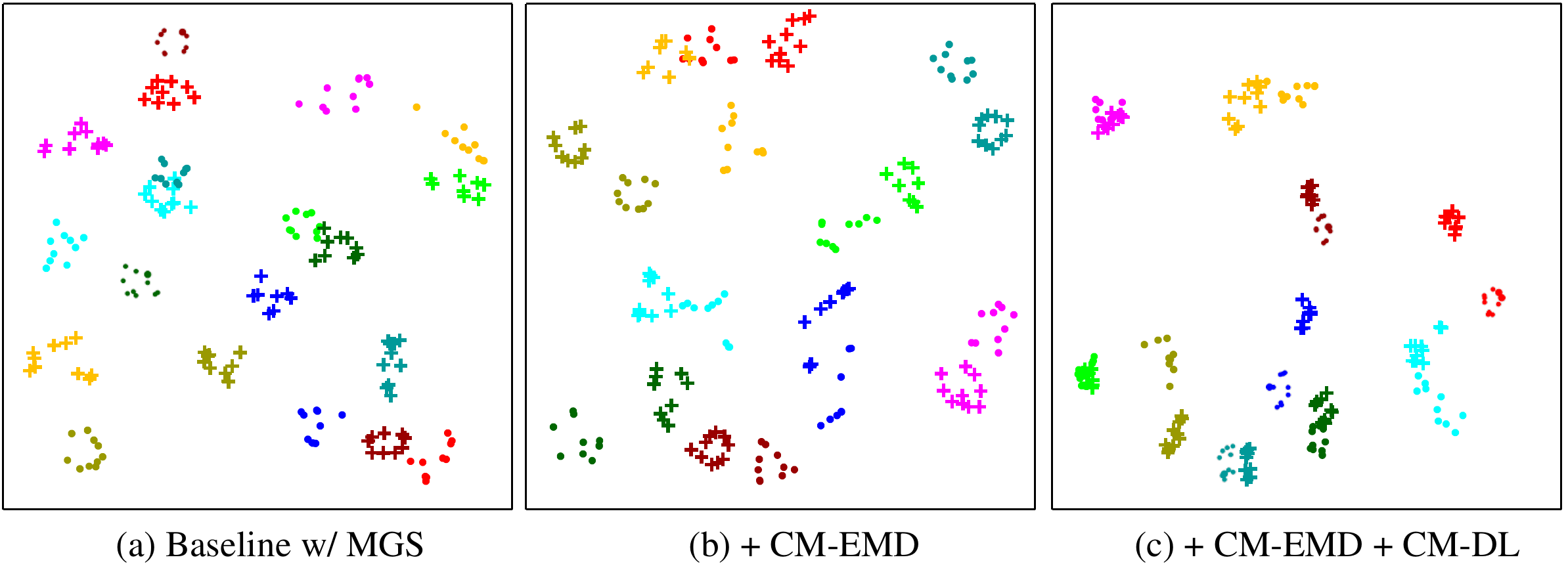}
  \caption{The t-SNE~\cite{maaten2008visualizing} of learned features for RegDB of (a) Baseline w/ MGS, (b) + CM-EMD, and (c) + CM-EMD + CM-DL. Colors represent the identities. Circle-dot denotes the visible modality and cross-mark represents the thermal modality. For better visualization, we randomly select 10 person identities with 10 samples for each modality from the testing set.}
  \label{fig tsne}
\end{figure*}

\subsection{Evaluation}

In this section, we conduct extensive experiments to investigate the effectiveness of the components of our model, \textit{i.e.}, MGS, CM-EMD, and CM-DL. Results are evaluated on the single-shot setting of all-search model for SYSU-MM01 and the ``Visible to Thermal'' setting for RegDB.

\textbf{Effectiveness of MGS}. To verify the advantage of the proposed Multi-Granularity Structure (MGS), we show the results of the baseline model with the global-based feature, local-based features and the multi-granularity features obtained by MGS, respectively. The comparisons are reported in Table~\ref{tab:ablation} (\#0 \textit{vs} \#2 \textit{vs} \#5). We can observe that using the local-based features can produce higher results than using the global-based features, especially on the RegDB dataset. In addition, using the multi-granularity features can further improve the results by a large margin, demonstrating the effectiveness of the proposed MGS.

\textbf{Effectiveness of accumulated local features}. We evaluate the Effectiveness of accumulated local features in Table~\ref{tab: acc}. We can find that accumulated local features can further improve the performance on two datasets, especially on RegDB.

\textbf{Effectiveness of CM-EMD}. We then evaluate the effectiveness of the proposed CM-EMD. From the results in Table~\ref{tab:ablation} (\#0 \textit{vs} \#1, \#2 \textit{vs} \#3, \#5 \textit{vs} \#6), we can find that the proposed CM-EMD can consistently improve the model performance by a large margin, no matter the types of features. For example, when using the multi-granularity features, the rank-1 accuracy of the baseline improves from 61.58\% to 71.26\% and 69.08\% to 92.86\% for SYSU-M001 and RegDB, respectively. These results verify the large effectiveness of our CM-EMD and show the compatibility between the proposed CM-EMD and MGS. In Table~\ref{tab:CM-EMD-Compare}, we compare CM-EMD with its variant and KL divergence. For the CM-EMD variant, we directly use cosine-similarities between samples to replace the weights obtained by the optimal transport strategy. The results show that CM-EMD largely outperforms the other two methods, further demonstrating the superiority of the proposed CM-EMD.
The advantage of CM-EMD over cosine-similarity based method is mainly due to that CM-EMD finds an overall optimal transport strategy to align two modalities while cosine-similarity based method only considers the relationship between individual pairs.

\begin{table}[!t]
  \centering
  \small
  \caption{Comparison of distribution alignment methods. KL: Kullback–Leibler divergence,  w/o OTS: use cosine-similarities to replace the weights obtained by optimal transport strategy.}
    \begin{tabular}{l|cc|cc}
    \hline
    \multirow{2}[1]{*}{Method} & \multicolumn{2}{c|}{SYSU-M001} & \multicolumn{2}{c}{RegDB} \\
\cline{2-5}          & R1    & mAP   & R1    & mAP  \\
    \hline
    Ours (w/o CM-EMD) + KL & 65.05  & 61.09  & 83.74 & 78.33 \\
    Ours w/o OTS & 69.18  & 65.36  & 90.05 & 82.40 \\
    Ours & \bf 73.39  & \bf 68.56  & \bf 94.37  & \bf 88.23 \\
    \hline
    \end{tabular}%
  \label{tab:CM-EMD-Compare}%
\end{table}%

\begin{table}[!t]
  \centering
  \small
  \caption{Comparison of CM-DL, center loss and triplet loss. Base: Baseline w/ MGF + CM-EMD}
    \begin{tabular}{l|cc|cc}
    \hline
    \multirow{2}[1]{*}{Method} & \multicolumn{2}{c|}{SYSU-M001} & \multicolumn{2}{c}{RegDB} \\
\cline{2-5}          & R1    & mAP   & R1    & mAP  \\
    \hline
    Base & 71.26  & 66.59  & 92.86 & 84.99 \\
    + Center loss~\cite{wen2016discriminative} & 69.94  & 64.44  & 91.41 & 83.44 \\
    + Triplet loss~\cite{hermans2017defense} & 71.21  & 66.39  & 91.02 & 81.83 \\
    + CM-DL & \bf 73.39  & \bf 68.56  & \bf 94.37  & \bf 88.23 \\
    \hline
    \end{tabular}%
  \label{tab: CM-DL-Compare}%
\end{table}%

\textbf{Effectiveness of CM-DL}. We next investigate the superiority of the proposed CM-DL. Since the loss of CM-DL is calculated on the holistic feature, which is generated based on the local-based features, we add CM-DL to models that include local-based features. The results in Table~\ref{tab:ablation} (\#3 \textit{vs} \#4, \#6 \textit{vs} \#7) show that injecting CM-DL into the model can obtain consistent improvements. This demonstrates the effectiveness of the proposed method and also validates the compatibility of the proposed three techniques (MGS, CM-EMD and CM-DL). In Table~\ref{tab: CM-DL-Compare}, we compare CM-DL with two popular metric learning methods, \textit{i.e.}, center loss~\cite{wen2016discriminative} and triplet loss~\cite{hermans2017defense}. We find that center and triplet losses can not improve the results but obtain a performance decrease on both SYSU-M001 and RegDB. However, our CM-DL achieves improvements on both datasets and clearly exceeds the other two methods. This further validates the benefits of our CM-DL in assisting CM-EMD. 
It is because that CM-DL is specifically designed to overcome the discrimination degradation problem caused by modality alignment, thus producing an improvement when adding to CM-EMD. However, given a model with a proper modality alignment, the constraints of center and triplet losses are mostly satisfied during training and thus will not bring further improvement without a more careful sampling strategy.

\textbf{Effectiveness of Trainable Weights}. For CM-DL, the holistic feature (Eq.~\ref{eq fh}) is obtained by concatenating local features with trainable weights. In Table~\ref{tab: pw}, we report the results of models with and without using trainable weights. When removing the trainable weights, we use equal weights for all parts. Table~\ref{tab: pw} shows that using trainable weights can consistently produce higher results on both datasets.

\begin{table}[!t]
  \centering
  \small
  \caption{Analysis of the trainable weights (TW) of CM-DL.}
    \begin{tabular}{l|cc|cc}
    \hline
    \multirow{2}[1]{*}{Method} & \multicolumn{2}{c|}{SYSU-M001} & \multicolumn{2}{c}{RegDB} \\
\cline{2-5}          & R1    & mAP   & R1    & mAP  \\
    \hline
    Ours w/o TW & 71.18  & 67.42  & 93.35 & 87.86 \\
    Ours & \bf 73.39  & \bf 68.56  & \bf 94.37  & \bf 88.23 \\
    \hline
    \end{tabular}%
  \label{tab: pw}%
\end{table}%

\subsection{Visualization}

To better reflect the superiority of the proposed methods, we visualize the top-10 retrieval results on the SYSU-M001 dataset. The compared methods are ``Baseline w/ MGS'', ``Baseline w/ MGS + CM-EMD'' and ``Baseline w/ MGS + CM-EMD + CM-DL''. The comparisons are illustrated in Fig.~\ref{fig retrieval}. It is clear that the proposed CM-EMD and CM-DL can significantly improve the ranking lists. That is, more positive samples are ranked in the top positions when using CM-EMD and CM-DL. 

\textbf{t-SNE of Feature Distribution.} We then show the t-SNE \cite{maaten2008visualizing} of features on the testing set for RegDB in Figure~\ref{fig tsne}. Results on the baseline and our proposed CM-EMD show that CM-EMD can effectively align the cross-modality gap. The CM-DL can further increase the similarity of the inter- and intra-modality features of the same identity.

\section{Conclusion}
\label{sec:conclusion}
In this paper, we propose a novel modality alignment method for visible thermal person re-identification, called Cross-Modality Earth Mover's Distance (CM-EMD), which can alleviate the impact of intra-identity variations and thus can achieve a more effective distribution alignment. Moreover, we introduce two methods to facilitate the benefit of the proposed CM-EMD, which are Cross-Modality Discrimination Learning (CM-DL) and Multi-Granularity Structure (MGS). CM-DL enables the model to learn more discriminative representation while MGS enables us to perform modality alignment in both coarse-grained to fine-grained levels. Experiments verify the advantages of the proposed methods. Our final solution can achieve the state-of-the-art results on SYSU-M001 and RegDB.

{\small
\bibliographystyle{IEEEtran}
\bibliography{ref}
}

\end{document}